\documentclass[runningheads]{llncs}
\usepackage{graphicx}

\usepackage{latexsym}
\usepackage{amsmath}
\usepackage{amsfonts}
\usepackage{booktabs}
\usepackage{multirow}
\usepackage{makecell}
\usepackage{algorithm}
\usepackage{algorithmic}
\usepackage{color}
\usepackage[misc]{ifsym}
\usepackage[marginal]{footmisc}

\begin{document}
\title{Knowledge-Grounded Dialogue with Reward-Driven Knowledge Selection}
%
%

\author{
Shilei Liu \and Xiaofeng Zhao \and Bochao Li \and  Feiliang Ren\thanks{Corresponding author.}
}

\authorrunning{S. Liu et al.}
\institute{School of Computer Science and Engineering\\ Northeastern University, Shenyang, 110169, China \\
\email{\{1901750, 1901840, 1901725\}@stu.neu.edu.cn\\ renfeiliang@cse.neu.edu.cn}}

\maketitle              

\begin{abstract}
    Knowledge-grounded dialogue is a task of generating a fluent and informative response based on both conversation context and a collection of external knowledge, in which knowledge selection plays an important role and attracts more and more research interest. However, most existing models either select only one knowledge or use all knowledge for responses generation. The former may lose valuable information in discarded knowledge, while the latter may bring a lot of noise. At the same time, many approaches need to train the knowledge selector with knowledge labels that indicate ground-truth knowledge, but these labels are difficult to obtain and require a large number of manual annotations. Motivated by these issues, we propose Knoformer, a dialogue response generation model based on reinforcement learning, which can automatically select one or more related knowledge from the knowledge pool and does not need knowledge labels during training. Knoformer is evaluated on two knowledge-guided conversation datasets, and achieves state-of-the-art performance.
\end{abstract}
\section{Introduction}
With the advances in sequence to sequence models,
neural dialogue systems have attracted more and more research attention.
Neural conversation response generation could be formulated as a Seq2Seq task: given a dialogue history, the model is asked to generate a high-quality response.
A large number of end-to-end generative neural conversation models have been applied to open-domain conversation and chatbot, have achieved success in generating fluent responses.
However, the usual Seq2Seq models tend to produce shorter and simpler responses, which are not informative~\cite{DBLP:conf/naacl/LiGBGD16,lian_learning_2019}.

In order to generate informative and meaningful responses, a number of methods have been proposed by leveraging external knowledge.
Besides the dialogue history, knowledge-based dialog system also combines several external knowledge (in this paper we only discuss unstructured textual knowledge) to construct an input sample. Generally, the collection of candidate knowledge is obtained through rough text retrieval in the knowledge base with a certain amount of noise. The noisy knowledge will lead the conversation to meaningless themes~\cite{lin_generating_2020}, so it is essential for a model to identify and select the appropriate knowledge.

However, most existing methods ~\cite{lian_learning_2019,dinan_wizard_2019,kim_sequential_2020,chen_bridging_2020} can only select one knowledge with the highest confidence from the candidate knowledge to participate in dialogue generation, while abandoning other knowledge with low confidence but possibly containing useful information. 
In order to make full use of the external knowledge, some approaches like ~\cite{lin_generating_2020} try to use all the knowledge to participate in the response generation. Nevertheless, when the size of candidate knowledge set is too large, this method will bring serious computational overheads and noise knowledge will lead the conversation to irrelevant topics.

Besides, many existing methods ~\cite{kim_sequential_2020,chen_bridging_2020} require labels that indicate the ground-truth knowledge to train the knowledge-grounded dialog models. However, these knowledge labels are difficult to obtain and need to be constructed through a large number of manual annotations, which is labor-intensive. 

Motivated by the above issues, we propose Knoformer, a novel knowledge-grounded dialogue response generation model. 
Firstly, we present a knowledge-aware dialogue module, which take the concatenation of dialogue history and selected knowledge as inputs and generation responses. 
We perform supervised learning to train this module. 
Secondly, we propose a knowledge selection module to select all appropriate knowledge according to hidden states from dialogue module. 
Due to the lack of knowledge labels, the selection module uses the feedback from the dialogue module as reward and uses the policy gradient algorithm for training. Besides, we also add a weak supervision loss to the selector to further boost the accuracy. 
The dialogue module and knowledge selector are optimized jointly in a recurrent way.

We conduct experiments on Wizard-of-Wikipedia~\cite{dinan_wizard_2019} and Holl-E~\cite{DBLP:conf/emnlp/MogheABK18}, and results shows that our model achieves the new state-of-the-art performance. 

\section{Related Work}
Knowledge-based conversation have shown promising results in improving response informativeness ~\cite{DBLP:conf/emnlp/ZhouPB18,dinan_wizard_2019}. 
PostKS~\cite{lian_learning_2019} uses prior and posterior distributions over knowledge to train a selector which can choice the most suitable candidate (and discard others) to participate in the response generation;
KIC~\cite{lin_generating_2020} uses recurrent knowledge interaction among response decoding steps incorporate  appropriate knowledge and uses pointer-generate networks copy tokens from external knowledge according to knowledge attention distribution;
~\cite{kim_sequential_2020} proposes Sequential Knowledge Transformer (SKT) to model knowledge selection history in multi-turn dialogue;
PIPM/KDBTS~\cite{chen_bridging_2020} improves on PostKS and SKT by using posterior information prediction and knowledge distillation to bridge the gap between prior and posterior knowledge selection.
PostKS, SKT and PIPM/KDBTS only select one knowledge but our Knoformer could select multiple knowledge from knowledge pool; To train SKT and PIPM/KDBTS, ground-truth knowledge labels are needed, but KnoFormer does not need to specify ground-truth knowledge in advance; KIC uses a special structure to realize knowledge attention which has poor portability, and has no ability to filter out noise knowledge, while our model can directly use Transformer to realize knowledge attention and has the ability to select highly relevant knowledge. 

One of the most related models to ours may be KnowledGPT~\cite{zhao_knowledge-grounded_2020}, who also focus on the multiple knowledge selection issue in the knowledge-grounded dialogue by reinforcement learning. Our work is novel in that during training, KnowledGPT only calculates one reward for all knowledge selection actions, which is equivalent to only one action in an \emph{episode}, while our model calculates rewards for each action, which can accurately punish or reward each action.

\section{Methodology}
We use capital letters for sequences (e.g., $Y$), lowercase letters for  tokens in sequences (e.g., $u$), bold capital letters for matrices (e.g., $\mathbf{K}$), and bold lowercase letters for vectors (e.g., $\mathbf{v}$).

Given a conversation history $U=\{u_1, u_2, ..., u_m\}$ and a set of external knowledge $\mathcal{K}=\{K_j\}_{j=1}^r$,
where $K_j=\{k_1^j, k_2^j, ..., k_n^j\}$ and $r$ is the size of external knowledge set,
the task of our proposed approach is to learn a knowledge selection module to select a subset of $\mathcal{K}$:
\begin{equation}
    p(\mathcal{K}_{sub}) = \prod_{i=1}^o p\left(K_{a_i} \mid U, K_{a_1},...,K_{a_{i-1}}\right) \quad \mathcal{K}_{sub} = \{K_{a_i}\}_{i=1}^o \quad 1 \leq a_i \leq r
\end{equation}
where $a_i$ represents the index of $i$-th knowledge in $\mathcal{K}_{sub}$. Specifically, we use a recurrent method to select knowledge one by one.
The first knowledge is selected based on the dialogue history $U$, and then the second knowledge is selected based on $U$ and the first knowledge $K_{a_1}$, and so on, to reach the maximum number of knowledge choices.
$\mathcal{K}$ usually has a lot of noise, so the filtered $\mathcal{K}$ is more closely related to the conversation topic.
In addition, we also learn a dialogue generation model generate response using $U$ and $\mathcal{K}_{sub}$: $p(\hat{Y}) = f(U, \mathcal{K}_{sub})$. 

Compared with some models~\cite{lian_learning_2019,kim_sequential_2020,chen_bridging_2020} that can only select one knowledge, our model uses a recurrent selection mechanism to pick multiple suitable external knowledge.

\begin{figure}[htbp]
    \centering
    \includegraphics[width=12cm]{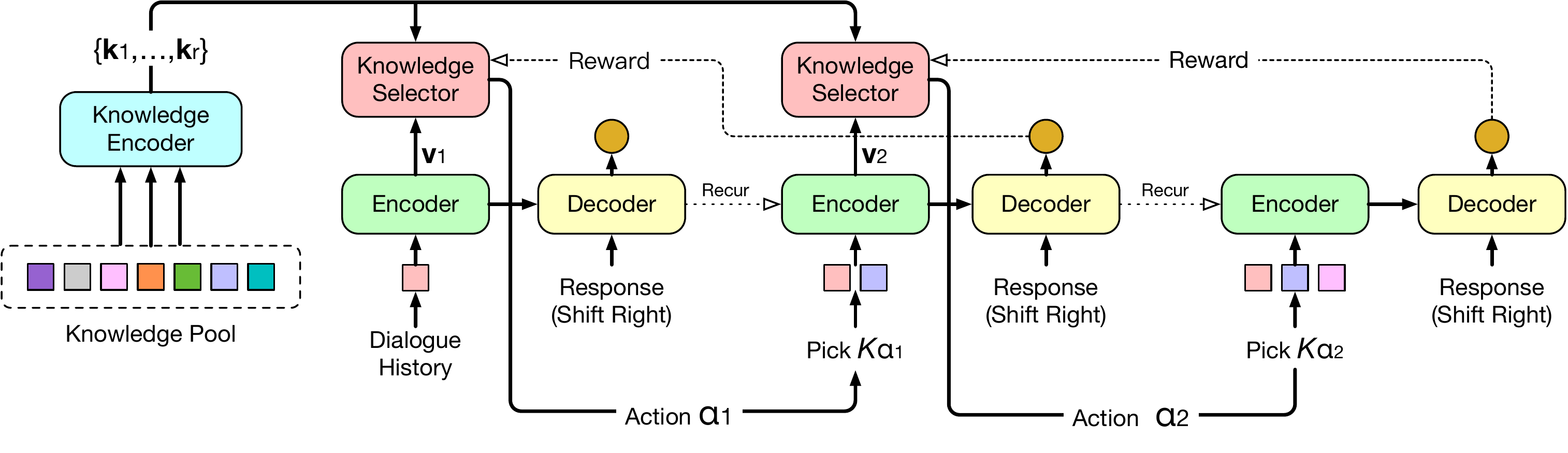}
    \caption{The architecture of our proposed model (suppose the model chooses two knowledge). Modules of the same color share parameters. The orange circles represent the loss of the dialogue model, and they will be accumulated and averaged to form the final dialogue loss $\mathcal{L}_u^\prime$.}
    \label{fig:struct}
\end{figure}

Our proposed model consists of three parts, a dialogue module which can fuse the features of knowledge and dialogue history to generate a response ($\S$ \ref{sec:diagmodel}), a knowledge encoding module which can encode all external knowledge separately ($\S$ \ref{sec:knomodel}), and a knowledge selection module which can be trained without ground-truth knowledge based on reinforcement learning ($\S$ \ref{sec:knoselmodel}).
$\S$ \ref{sec:traininfer} introduces the training and inference of our model.
The architecture of our proposed model is show in Fig. \ref{fig:struct}.
\subsection{Dialogue Module}\label{sec:diagmodel}
Our model operates in an iterative manner. Each time a knowledge is selected, the dialogue module will perform feature fusion on the currently selected knowledge. Our dialogue module is based on Transformer~\cite{vaswani_attention_2017}, which takes the dialogue history and several selected external knowledge as input, and generates a response token by token.

Assuming that $c-1$ knowledge has been selected at present, then the input of the dialogue model is $X_c=\left[\texttt{[SOS]};U;K_{a_1};...;K_{a_{c-1}};\texttt{[EOS]}\right]$, where \texttt{[SOS]} and \texttt{[EOS]} are special tokens.
We pass $X_c$ to Transformer encoder to obtain the context-aware representation $\mathbf{X}_c$:
\begin{equation}
    \mathbf{X}_c = \text{TFEncoder}\left(e(X_c), \theta_e \right) \in \mathbb{R}^{m \times d}
    \label{eq:encodeutt}
\end{equation}
where $e(\cdot)$ means to convert a sequence (or token) into word embedding vectors using a lookup table $\mathbf{M} \in \mathbb{R}^{vocab \times d}$ and $\theta_e$ represents the learnable parameters. After encoding, the information of dialogue history and external knowledge is fully integrated.

We take $\mathbf{X}_c$ (as memory) and golden response $Y$ (as input) into Transformer decoder to generate response token by token. 
\begin{equation}
    \mathbf{h}_t = \text{TFDecoder}\left(\mathbf{X}_c, e(Y_{1:t-1}),\theta_d \right) \in \mathbb{R}^{d} \quad p(y_t) = \frac{e(y_t)\mathbf{h}_t^\top}{\sum_{y^\prime} e(y^\prime)\mathbf{h}_t^\top}
\end{equation}

\subsection{Knowledge Encoder}\label{sec:knomodel}
For $i$-th ($1 \leq i \leq r$) knowledge $K_i$ in external knowledge set, we wrap it with two special characters \texttt{[SOS]} and \texttt{[EOS]}, and sent to a Transformer encoder:
\begin{equation}
    \mathbf{K}_i = \text{TFEncoder}\left(e(K_i), \theta_k \right) \in \mathbb{R}^{n \times d}
    \label{eq:encodekno}
\end{equation}
We use the representation of \texttt{[SOS]} token (denoted as $\mathbf{k}_i \in \mathbb{R}^{d}$) as the overall representation of $i$-th knowledge.

\subsection{Knowledge Selector}\label{sec:knoselmodel}

Assuming that we will choose the $c$-th knowledge in current step, we formulate the problem of learning-to-select under the framework of reinforcement learning. We define the \textbf{state} $s=\{K_{a_1}, ... , K_{a_{c-1}}\}$ (omit some constants) of the model to be the knowledge that the selector has selected before the current step, e.g., $s=\{3, 2, 1, 4\}$.
The \textbf{action} $a$ is which knowledge to select.
We define the \textbf{action space} of current step as the knowledge index in $\mathcal{K}$ (i.e., \{1,2,...,r\}). During implementation, we use mask technology  to avoid choosing repeated action. 
Since $\mathbf{X}_c$ integrates the features from dialogue history and selected knowledge, in order to propagate the information across different steps, we represent the state $s$ with the representation of \texttt{[SOS]} token (denoted by $\mathbf{v}_c \in \mathbb{R}^d$) in $\mathbf{X}_c$.

The selection policy gives the probability $\pi \left(a \mid s\right)$ of taking an action $a$ at the current state $s$, which is modeled by a bilinear matrix:
\begin{equation}
    \pi \left(a \mid s\right) = \mathop{softmax}_{a \in \mathcal{A}_c}\left(\mathbf{v}_c \mathbf{W} \mathbf{k}_a^\top\right)
    \label{eq:act}
\end{equation}
where $\mathbf{W} \in \mathbb{R}^{d \times d}$ is a learnable parameter and $\mathbf{k}_a$ is the knowledge representation ($\S$ \ref{sec:knomodel}). In order to achieve the early stop mechanism, a stop action can be added to action space in implementation.
\subsection{Modules Integration} \label{sec:traininfer}

In the training phase of the recurrent selection mechanism, the actions of selecting next knowledge are sampled according to the probability given by the selection policy. Our model generates a sequence of knowledge for each dialogue history. We train the dialogue model with supervised learning, and we train the selector network via reinforcement learning and weekly supervised learning.

\subsubsection{Supervised Learning for Response Generation} The ground-truth response has been given, so we train the conversation model via supervised learning. Consistent with most Seq2Seq models, the training objective is to minimize the negative log likelihood (NLL):
\begin{equation}
    \mathcal{L}_u = -\frac{1}{z}\sum_{t=1}^z \log p (y_t)
    \label{eq:diagloss}
\end{equation}

\subsubsection{Reinforcement Learning for Selection Policy} Based on the assumption that the ground-truth knowledge labels are not available, supervised learning cannot be used to train the knowledge selection model, it is natural to train it via reinforcement learning.

First of all, the accumulated reward for taking action $a$ at state $s$ is denoted as $\text{R}(s, a)$, which is derived in a recursive manner:
\begin{equation}
    \text{R}(s, a) =  e^{-\text{PPL}_{(s,a)} / \gamma} + \text{R}(s^\prime, a^\prime)
\end{equation}
where $\gamma$ is a constant and we empirically set it to $10$. $\text{PPL}_{(s,a)}$ denotes the perplexity
of ground-truth response $Y$ in current step and $\text{R}(s^\prime, a^\prime)$ denotes the next state-action pair. We use perplexity as reward to make the knowledge selector more inclined to select knowledge that can generate high-quality responses.

The selection policy network can be trained by maximizing the expected accumulated reward through the policy gradient algorithm ~\cite{DBLP:journals/ml/Williams92}:
\begin{equation}
    \mathcal{J} = \mathbb{E}_{a\sim \pi}\left[\tilde{\text{R}}(s, a)\right]
    \label{eq:accreward}
\end{equation}
where $\tilde{\text{R}}(s, a)=\text{R}(s, a)-b$ and $b \approx \mathbb{E}[\text{R}(s, a)]$ is the baseline that is
used to reduce the variance of gradient estimation~\cite{DBLP:conf/emnlp/ClarkM16}.
To be consistent with the notations in response generation, we denote the loss function of selection policy as $\mathcal{L}_k$, which is the negative expected accumulated reward $\mathcal{J}$ in Eq. \ref{eq:accreward}: $\mathcal{L}_k = -\mathcal{J}$. Thus, the gradient of $\mathcal{L}_k$ over a series of action-pair $\mathcal{B}$ is given by:
\begin{equation}
    \nabla \mathcal{L}_k = - \sum_{(s,a)\in \mathcal{B}}\nabla \log \pi \left(a \mid s\right) \text{R}(s, a)
\end{equation}
\subsubsection{Weakly Supervised Learning for Selector}
Valuable knowledge usually has a higher text similarity with the ground-truth response, so we add an additional selection loss to the selector when selecting the first knowledge. We take the response as the query, use the TF-IDF algorithm to score the knowledge in $\mathcal{K}$, mark the index of knowledge with the highest score as $a^+$, and the additional loss is defined as:
\begin{equation}
    \mathcal{L}_s = -\log \pi \left(a^+ \mid s\right)
    \label{eq:knoloss}
\end{equation}

\subsubsection{Training and Inference} The training procedure is show in Algorithm \ref{alg:rlconv} (for ease of understanding, assume batch size is 1 and only optimize one step).
The overall loss is defined as $\mathcal{L} = \mathcal{L}_u^\prime /o+ \mathcal{L}_k + \lambda \mathcal{L}_s$. Weakly supervised labels may not be correct, so we put a smaller weight $\lambda$ on $\mathcal{L}_s$.

\begin{figure}[htbp]
    \centering
    \includegraphics[width=7.5cm]{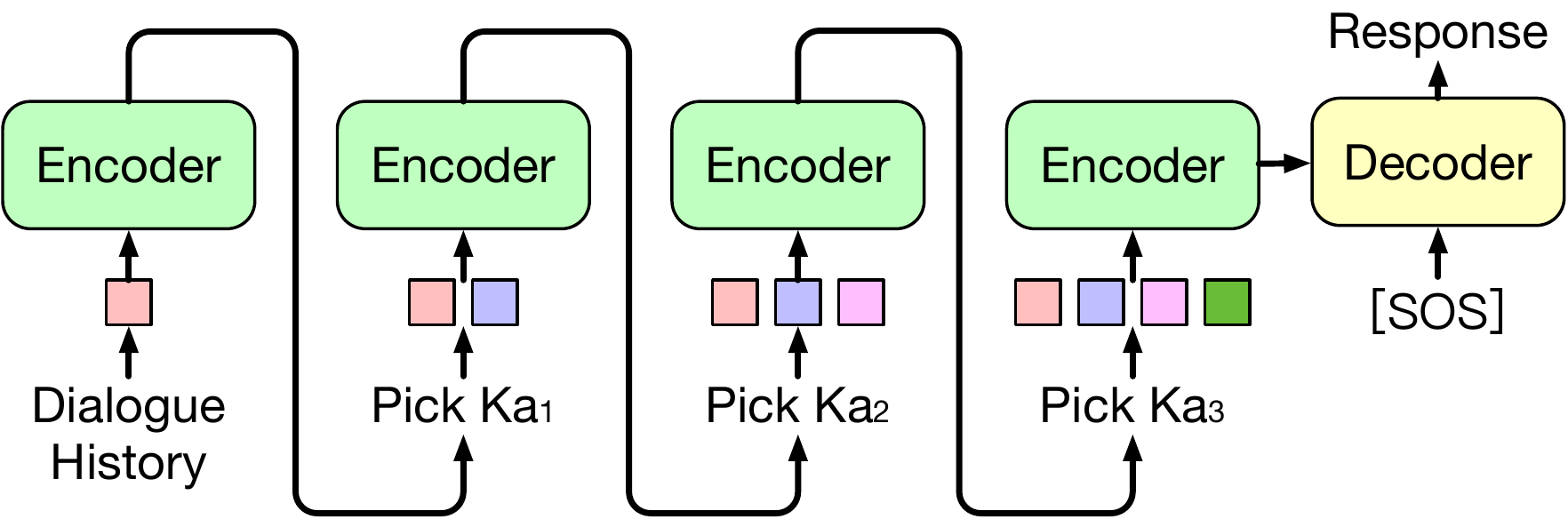}
    \caption{Sketch of inference.}
    \label{fig:struct_eval}
\end{figure}
During inference, the knowledge selector take the best action (instead of sampling) at step $c$ according to the selection policy:
\begin{equation}
    a^* = \mathop{\arg\max}_{a \in \mathcal{A}_c} \pi  \left(a \mid s\right)
\end{equation}
After the action $a^*$ is taken, a new knowledge is taken from the knowledge pool and appended to the input sequence, and so on, until reaching the upper bound of knowledge selection $o$. Because there is no need to calculate the reward during inference, in order to save overhead, the decoder only decodes when $c=o$. The sketch of inference is shown in Fig  \ref{fig:struct_eval}.

\begin{algorithm}[tb]
\small
    \caption{Optimization Step}
    \label{alg:rlconv}
    \textbf{Input}: $U$, $Y$,  $\mathcal{K}=\{K_i\}_{i=1}^r$ and number of selections $o$;\\
    \textbf{Output}: Generation loss $\mathcal{L}_u^\prime$, selection loss $\mathcal{L}_k$ and $\mathcal{L}_s$;
    \begin{algorithmic}[1] 
        \STATE Encode knowledge independently according to Eq. \ref{eq:encodekno} and get the knowledge representation $\{\mathbf{k}_i\}_{i=1}^r$;
        \STATE Initialize $s=\phi$, $\mathcal{B}=\phi$, $\mathcal{L}_u^\prime=0$;
        \FOR{$c$ \textbf{in} $\{1,...,o\}$}
        \STATE Construct $X_c$ from U and $s$ according to $\S$ \ref{sec:diagmodel};
        \STATE Encode $X_c$ to obtain $\mathbf{X}_c$ and $\mathbf{v}_c$ according to Eq. \ref{eq:encodeutt};
        \STATE Decode and calculate $\mathcal{L}_u$ according to Eq. \ref{eq:diagloss};
        \STATE $\mathcal{L}_u^\prime \gets \mathcal{L}_u^\prime + \mathcal{L}_u$;
        \IF {$c=1$}
        \STATE Calculate $\mathcal{L}_s$ according to Eq. \ref{eq:knoloss};
        \ENDIF
        \IF {$c<o$}
        \STATE Calculate $\pi$ according to Eq. \ref{eq:act};
        \STATE Sample an action $a$ using Gumbel-Max Trick;
        \STATE $\mathcal{B} \gets \mathcal{B} \cap \{(s, a)\}$, $s \gets s \cap \{a\}$;
        \ENDIF
        \ENDFOR
        \STATE Calculate $\mathcal{L}_k$ according to Eq. \ref{eq:accreward};
        \STATE \textbf{return} $\mathcal{L}_u^\prime$, $\mathcal{L}_s$ and $\mathcal{L}_k$;
    \end{algorithmic}
\end{algorithm}

\section{Experiments}
\subsection{Datasets}
We conduct experiments on Wizard-of-Wikipedia~\cite{dinan_wizard_2019} and Holl-E~\cite{DBLP:conf/emnlp/MogheABK18}. Wizard-of-Wikipedia contains 18,430 training dialogues, 1,948 validation dialogues and 1,933 test dialogues on 1365 topics.  And test set is split into two subsets according to topics, which are Test Seen with 965 dialogues and Test Unseen with 968 dialogues whose topics are never seen in training and validation set. There are about 61 sentences on average in the knowledge pool per turn, which are retrieved from Wikipedia based on the context.
Holl-E contains 7,228 training dialogues, 930 validation dialogues and 913 test dialogues. Each dialogue is assigned with about 60 documents on average as the knowledge pool. Here, we use the modified version ~\cite{kim_sequential_2020} which fits for knowledge selection.

It should be noted that in this paper, we focus on the scenarios where ground-truth knowledge is unknown. Thus, we does not use the ground-truth knowledge labels.

\subsection{Models for Comparison}
We compare our Knoformer with the following baselines:
\begin{itemize}
    \item \textbf{KnowledGPT}: Another knowledge-based dialogue system based on BERT and GPT-2 which can select multiple external knowledge via reinforcement learning~\cite{zhao_knowledge-grounded_2020}. KnowledGPT does not provide complete source code and experimental results on other datasets, so we can only compare with it on the Wizard-of-Wikipedia dataset.
    \item \textbf{TF-IDF}: TF-IDF is a commonly used algorithm in information retrieval. We perform TF-IDF to sort all documents in knowledge set (dialogue context as query), and concatenate dialogue context and top-$20$ of them as the input.
    \item \textbf{PIPM/KDBTS}: A latent variable model that uses specially designed Posterior Information Prediction Module (PIPM) to select knowledge and Knowledge Distillation Based Training Strategy (KDBTS) to train the decoder with the knowledge selected from the prior distribution~\cite{chen_bridging_2020}. It is an improvement of PostKS and SKT, but still can only select one knowledge.
\end{itemize}
For fair comparison, we only use the knowledge selector of the above methods, and use BART~\cite{lewis_bart_2020} as the unified dialogue response generator. It should be noted that most pre-trained language models (e.g., BERT, GPT-2, BART) has a limit of max number of input tokens they can handle, so for BART+TF-IDF, text exceeding the maximum length will be truncated, and only keep the first 384 tokens. The source code of KIC and some other methods is not available, so we will not compare with them. 

\subsection{Implement Details}
We implement our model over PyTorch framework.
The parameters of dialogue module and knowledge encoder are initialized with BART-base and BERT-base respectively.
We train our model using AdamW \cite{DBLP:journals/corr/abs-1711-05101} optimizer with a batch size of 16 and learning rate 5e-5 at 3 epochs on a NVIDIA QUADRO RTX 8000 machine, and other hyperparameters are detailed in Table \ref{tab:hyper}.
When decoding, the number of beams is between 1 to 5.
We adopt widely used public evaluation toolkit NLTK to evaluate the model performance.
\begin{table}[htbp]
    \caption{Hyperparameters over different datasets.}
    \small
    \centering
    \setlength{\tabcolsep}{2mm}{
        \begin{tabular}{lccc}
            \toprule
                                    & Wizard seen & Wizard Unseen & Holl-E \\ \midrule
            Max Context Length      & 384         & 384           & 256    \\
            Max Knowledge Length    & 64          & 64            & 48     \\
            Max Response Length     & 48          & 48            & 48     \\
            Max Number of Selection & 6           & 3             & 2      \\ 
            $\lambda$ & 0.3           & 0.7             & 0.7      \\
            \bottomrule
        \end{tabular}}
    \label{tab:hyper}
\end{table}

\subsection{Automatic Evaluation}
We adopt three automatic metrics include BLEU, Div, and corpus-level unigram F1 for evaluation. BLEU is well-known machine translation evaluation metric, and generated response with higher BLEU / F1 is closer to the ground-truth response and has preferable fluency. Div-$n$ 
reflects the $n$-gram diversity of text \cite{DBLP:conf/naacl/LiGBGD16}, and 
 response with higher Div-$n$  could present more information.
\begin{table}[htbp]
    \caption{Automatic evaluation results on Wizard-of-Wikipedia test seen. BART in the first row means no knowledge is used.}
    \centering
    \small
    \setlength{\tabcolsep}{2mm}{
        \begin{tabular}{lcccccc}
            \toprule
            Model           & F1   & BLEU-1 & BLEU-2 & BLEU-3 & Div-1 & Div-2 \\ \midrule
            BART            & 24.0 & 22.4   & 10.2   & 5.0    & 6.0   & 23.8  \\
            BART+KnowledGPT & 26.2 & 25.6   & 13.6   & 8.8    & 7.8   & 24.6  \\
            BART+TF-IDF     & 24.8 & 23.3   & 11.0   & 5.9    & 7.1   & 29.8  \\
            BART+PIPM       & 25.0 & 23.6   & 11.3   & 6.1    & 7.0   & 28.7  \\\midrule
            Knoformer       & \textbf{27.4} & \textbf{26.4}   & \textbf{14.3}   & \textbf{9.2}    & \textbf{7.9}   & \textbf{32.1}  \\ \bottomrule
        \end{tabular}}
    \label{tab:wowseen}
\end{table}

\begin{table}[htbp]
    \caption{Automatic evaluation results on Wizard-of-Wikipedia test unseen.}
    \centering
    \small
    \setlength{\tabcolsep}{2mm}{
        \begin{tabular}{lcccccc}
            \toprule
            Model           & F1   & BLEU-1 & BLEU-2 & BLEU-3 & Div-1 & Div-2 \\ \midrule
            BART            & 23.1 & 21.6   & 9.5    & 4.7    & 4.5   & 19.6  \\
            BART+KnowledGPT & 24.7 & 24.6   & \textbf{12.6}   & 7.8    & 4.9   & \textbf{23.6}  \\
            BART+TF-IDF     & 23.5 & 22.3   & 10.0   & 5.4    & 5.1   & 21.4  \\
            BART+PIPM       & 24.0 & 23.0   & 10.5   & 5.6    & 4.7   & 20.8  \\\midrule
            Knoformer       & \textbf{25.4} & \textbf{24.8}   & \textbf{12.6}   & \textbf{8.0}    & \textbf{5.1}   & 23.1  \\ \bottomrule
        \end{tabular}}
    \label{tab:wowunseen}
\end{table}
Table \ref{tab:wowseen} and Table \ref{tab:wowunseen} shows the automatic evaluation results on Wizard, and Table \ref{tab:holle} shows the results on Holl-E.
We have the following observations:
(1) Our Knoformer significantly surpasses all baselines in most evaluation metrics of all datasets, which means our knowledge selection module is more targeted and can select more valuable knowledge;
(2) External knowledge is of great help to improve performance. If external knowledge is removed, performance will decline; 
(3) Only using prior experience (TF-IDF with context as query) to select knowledge is not effective.

\begin{table}[htbp]
    \caption{Automatic evaluation results on Holl-E.}
    \centering
    \small
    \setlength{\tabcolsep}{2mm}{
        \begin{tabular}{lcccccc}
            \toprule
            Model       & F1   & BLEU-1 & BLEU-2 & BLEU-3 & Div-1 & Div-2 \\ \midrule
            BART        & 24.7 & 20.6   & 11.5   & 5.2    & 4.0   & 16.1  \\
            BART+TF-IDF & 27.8 & 25.0   & 11.0   & 5.9    & \textbf{7.1}   & \textbf{28.3}  \\
            BART+PIPM   & 36.3 & 33.4   & 27.5   & 24.6   & 4.6   & 20.8  \\\midrule
            Knoformer   & \textbf{39.5} & \textbf{37.9}   & \textbf{31.0}   & \textbf{28.4}   & 7.0   & 25.8  \\ \bottomrule
        \end{tabular}}
    \label{tab:holle}
\end{table}
\subsection{Human Evaluation}
Besides automatic evaluation, we also recruit three human annotators to do qualitative analysis on response quality. For each corpus, we randomly sample 200 samples, and each sample contains the dialog history, response, and external knowledge set. The annotators then judge the quality of the responses from three aspects, including context coherence, language fluency and response diversity, and assign a score in \{0, 1, 2\} to each response for each aspect. Each response receives 3 scores per aspect, and the agreement among the annotators is measured via Fleiss’ kappa~\cite{fkappa}. The human evaluation result is shown in Table \ref{tab:humaneval}, and we observe that responses from our Knoformer are more fluent and more contextually coherent than those from baselines.
\begin{table}[htbp]
    \caption{Human evaluation results on Wizard-of-Wikipedia. }
    \centering
    \small
    \setlength{\tabcolsep}{1.8mm}{
        \begin{tabular}{lcccccccc}
            \toprule
            \multirow{2}{*}{Models} & \multicolumn{4}{c}{Wizard Test Seen} & \multicolumn{4}{c}{Wizard Test Unseen}                                             \\ \cmidrule(lr){2-5}\cmidrule(lr){6-9}
                                    & CC                                   & LF                                     & RD   & Kappa & CC   & LF   & RD   & Kappa \\ \midrule
            BART+KnowledGPT         & 1.78                                 & 1.80                                   & 1.64 & 0.61  & 1.72 & 1.74 & 1.66 & 0.62  \\
            BART+PIPM               & 1.73                                 & 1.75                                   & 1.65 & 0.59  & 1.69 & 1.70 & 1.62 & 0.61  \\
            Ours                    & 1.80                                 & 1.84                                   & 1.69 & 0.60  & 1.74 & 1.77 & 1.66 & 0.61  \\\bottomrule
        \end{tabular}}
    \label{tab:humaneval}
\end{table}
\subsection{Analysis}

\subsubsection{Ablation study.} In order to explore the importance of components in Knoformer, we conducted ablation experiments on Wizard-of-Wikipedia valid set, and the results are summarized in Table \ref{tab:as}. 
First of all, we remove the loss item $\mathcal{L}_k$ and $\mathcal{L}_s$ of knowledge selection, making the selection of knowledge completely random. Results show that meaningless selection will harm the performance of the model. 
Secondly, we remove $\mathcal{L}_s$ only, and performance is also degraded, which means that valuable knowledge does have a higher similarity with the ground-truth response, and building the link between knowledge and response directly helps improve model performance.  
Besides, we change the joint training to separate training, which means that selector and dialogue module are trained alternately. Results indicate that joint training can make better use of the feedback of dialogue model. 
In addition, when replacing our reward function with that of KnowledGPT (only return a reward for all actions in an episode), the performance drops significantly, which means that gives a reward to each action can effectively punish bad actions and enhance the knowledge selection. 

\begin{table}[htbp]
    \caption{Results on Wizard valid set.}
    \centering
    \small
    \setlength{\tabcolsep}{2.2mm}{
        \begin{tabular}{lcccccc}
            \toprule
            \multirow{2}{*}{Models} & \multicolumn{3}{c}{Wizard Test Seen} & \multicolumn{3}{c}{Wizard Test Unseen}                                   \\ \cmidrule(lr){2-4}\cmidrule(lr){5-7}
                                    & F1                                   & BLEU-1                                 & BLEU-3 & F1   & BLEU-1 & BLEU-3 \\ \midrule
            Ours                    & 27.1                                 & 26.5                                  & 9.2   & 25.3 & 24.6   & 8.2   \\
            -$\mathcal{L}_k,\mathcal{L}_s$        & 24.6                                 & 23.1                                   & 5.9   & 23.6 & 22.1   & 5.3   \\
            -$\mathcal{L}_s$        & 26.8                                 & 26.3                                   & 8.5   & 24.9 & 24.1   & 7.6   \\
            -joint                  & 26.9                                 & 26.3                                   & 9.1   & 25.1 & 24.3   & 8.0   \\
            $\text{R}^\dagger$       & 26.2                                 & 25.7                                   & 8.2   & 24.1 & 23.4   & 7.3   \\ \bottomrule
        \end{tabular}}
    \label{tab:as}
\end{table}

\subsubsection{Case study.} Table \ref{tab:case1} shows a case in Wizard test unseen. Our model select two pieces of knowledge and incorporate them into response. Compared with baselines and even reference, response generated by Knoformer is more informative. 
\begin{table}[htbp]
\scriptsize
\centering
\caption{A case from test unseen of Wizard-of-Wikipedia.}
\begin{tabular}{ll}
\toprule
\begin{tabular}[c]{@{}l@{}}Context\end{tabular}    & \begin{tabular}[c]{@{}l@{}}A: Elvis was such an amazing singer, but he also an incredible musician and actor. \\B: Yes, The King. Did you know he sold more than 1 billion units in 20 years.\end{tabular}
        \\ \midrule
 Reference    & \begin{tabular}[c]{@{}l@{}}I didn't know that! He was born in 1935 and died in 1977 though.\end{tabular}           \\ \midrule
\begin{tabular}[c]{@{}l@{}} Selected\\Knowl.\end{tabular}       & \begin{tabular}[c]{@{}l@{}}1. Elvis Aaron Presley (January 8, 1935 – August 16, 1977) was an American singer... \\ 2. ... he is often referred to as the King of Rock and Roll or simply the King. 
\end{tabular}
        \\ \midrule
\multicolumn{2}{l}{\begin{tabular}[c]{@{}l@{}}
\textbf{(Ours)} Yes, he was king of rock and roll. He was born in 1935 and died in 1977.  \\
\textbf{($\text{BART}$+TF-IDF)} Yes, he sold over 100 million records worldwide. \\
\textbf{($\text{BART}$+PIPM)}  I did not know that.  I do know that he founded the Wall Street firm in 1960.  
\end{tabular}}                                                                                                                                            \\ \bottomrule
\end{tabular}
\label{tab:case1}
\end{table}
\subsubsection{Impact of $o$.} 
\begin{figure}[htbp]
    \center
    \includegraphics[width=12cm]{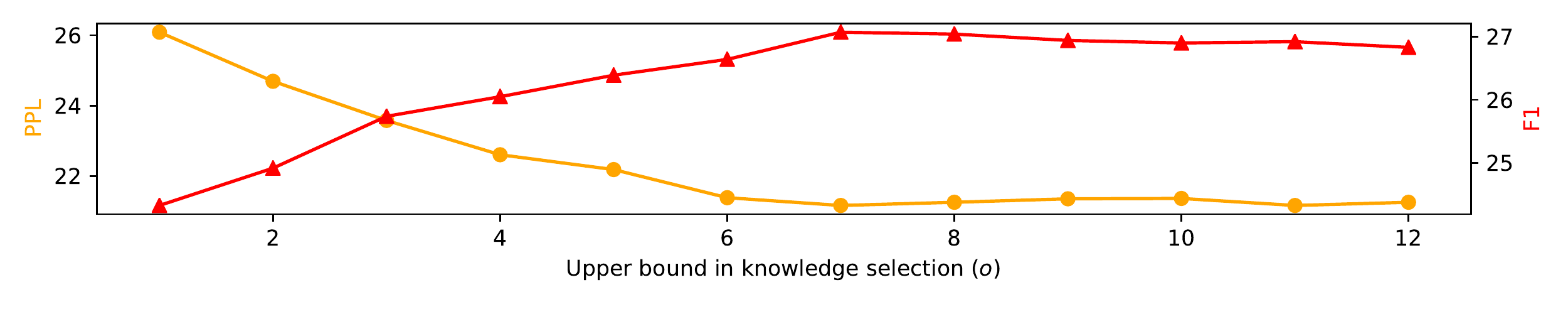}
    \caption{The performance of the model with different $o$ in Wizard valid set (seen).}
    \label{fig:io}
\end{figure}

To explore the influence of the number of knowledge choices on model performance, we vary the value of $o$ in $\{1, 2, ..., 12\}$ and report the evaluation results in Fig \ref{fig:io}. The smaller $o$, the smaller the probability that ground-truth knowledge will be captured. When $o$ reaches a certain extent, the performance improvement is very weak or even slightly decreased, implying that the noise in the knowledge pool will interfere with the generation of responses.

\section{Conclusion}
In this paper, we propose Knoformer, a knowledge-grounded dialogue generation model.
Evaluation results on four benchmarks indicate that our model can significantly outperform state-of-the-art methods. 
\section*{Acknowledgments}
This work is supported by the National Key R\&D Program of China (No.2018YFC0830701), the National Natural Science Foundation of China (No.61572120), the Fundamental Research Funds for the Central Universities (No.N181602013 and No.N171602003), Ten Thousand Talent Program (No.ZX20200035), and Liaoning Distinguished Professor (No.XLYC1902057).
%
%
%
\bibliographystyle{splncs04}
\bibliography{ijcai21}
\end{document}